\documentclass[journal]{IEEEtran}
\ifCLASSINFOpdf
\else
\fi

\usepackage{times}
\usepackage{float}
\usepackage{epsfig}
\usepackage{graphicx}
\usepackage{amsmath}
\usepackage{amssymb}
\usepackage{array}
\usepackage{mathabx}
\usepackage{threeparttable}
\usepackage{color}
\usepackage{diagbox}
\usepackage{multirow}
\usepackage{amsfonts,fancyhdr}

\begin{document}
%
\title{Learning Multi-level Deep Representations for Image Emotion Classification}
%
%
%

\author{Tianrong~Rao,~\IEEEmembership{Student Member,~IEEE,}
        Min~Xu,~\IEEEmembership{Member,~IEEE,}
        Dong~Xu,~\IEEEmembership{Senior Member ,~IEEE,}
\thanks{T. Rao and M. Xu are with School of Computing and Communications,University of Technology, Sydney, NSW, 2007, Email: Min.Xu@uts.edu.au}
\thanks{Dong Xu is with School of Electrical and Information Engineering, University of Sydney, NSW, 2006, Australia.}}

%
%

\markboth{Journal of \LaTeX\ Class Files,~Vol.~13, No.~9, September~2014}%
{Shell \MakeLowercase{\textit{et al.}}: Bare Demo of IEEEtran.cls for Journals}
%



\maketitle

\begin{abstract}
In this paper, we propose a new deep network that learns multi-level deep representations for image emotion classification (MldrNet). Image emotion can be recognized through image semantics, image aesthetics and low-level visual features from both global and local views. Existing image emotion classification works using hand-crafted features or deep features mainly focus on either low-level visual features or semantic-level image representations without taking all factors into consideration. The proposed MldrNet combines deep representations of different levels, i.e. image semantics, image aesthetics and low-level visual features to effectively classify the emotion types of different kinds of images, such as abstract paintings and web images. Extensive experiments on both Internet images and abstract paintings demonstrate the proposed method outperforms the state-of-the-art methods using deep features or hand-crafted features. The proposed approach also outperforms the state-of-the-art methods with at least 6\% performance improvement in terms of overall classification accuracy.
\end{abstract}


%
\IEEEpeerreviewmaketitle

\section{Introduction}
%
%
%
%
\IEEEPARstart {P}{sychological} studies have already demonstrated that humans' emotion reflections vary with different visual stimuli, such as images and videos \cite{lang1979bio,joshi2011aesthetics}. Inspired by these studies, computer scientists began to predict the emotional reactions of people given a series of visual contents. This creates a new research topic, called affective image analysis, which attracts increasing attention in recent years \cite{machajdik2010affective,zhao2014exploring,rao2016multi,zhao2016continuous}. However, compared to semantic-level image analysis, analyzing images at affective-level is more difficult, due to the two challenges of the complexity and subjectivity of emotions \cite{wei2004image}.

\begin{figure}
    \begin{center}
    \includegraphics[width=1\linewidth]{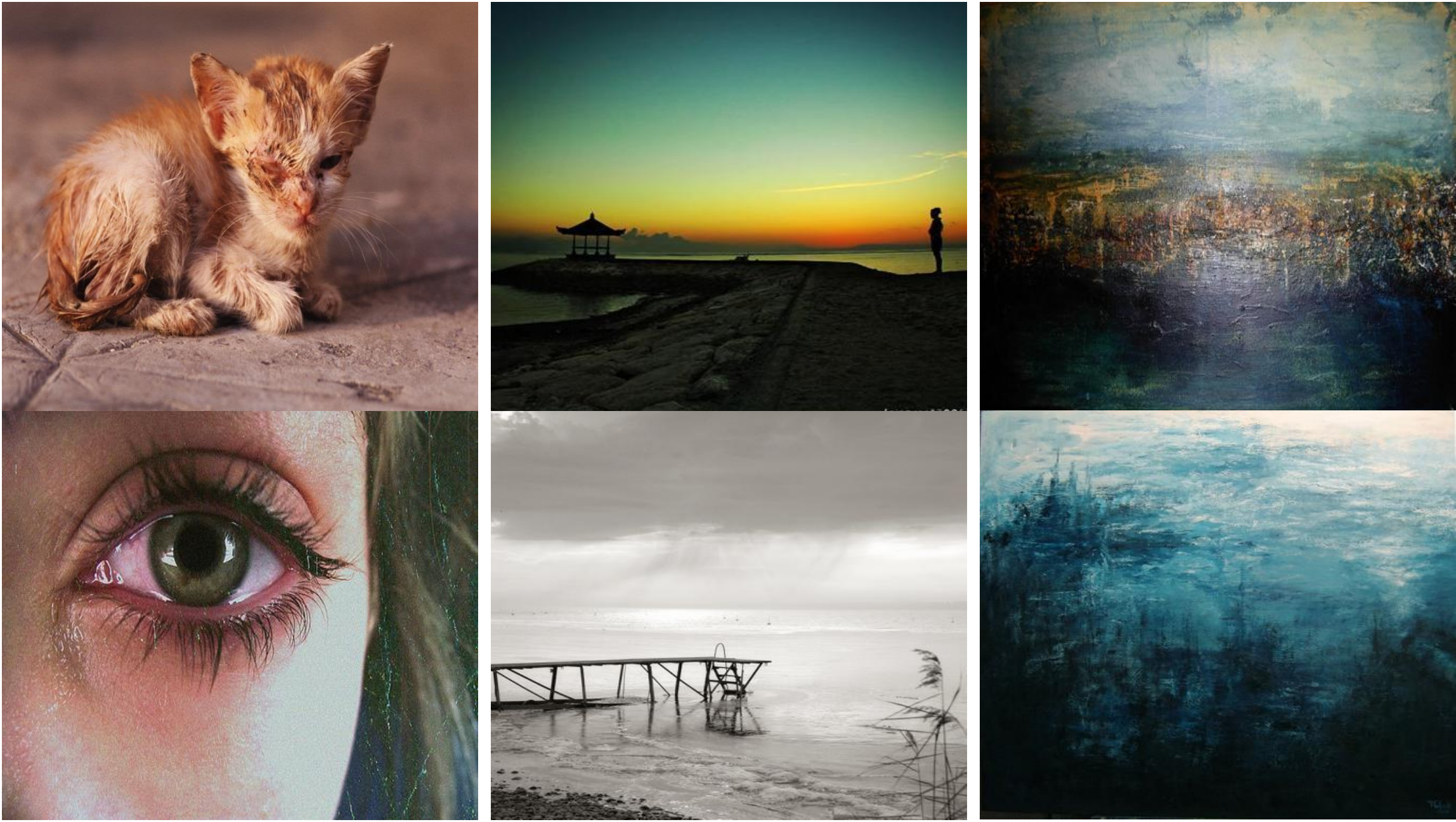}
    \end{center}
    \caption{Sample images from different datasets that evoke the same emotion \emph{sadness}. We can find out that image emotion is related to many factors. Left: web images whose emotions are mainly related to image semantics. Middle: art photos whose emotions are mainly related to image aesthetics, such as compositions and emphasis. Right: abstract paintings whose emotions are mainly related to low-level visual features, such as texture and color.}
    \label{fig:sadsample}
\end{figure}

As shown in Figure \ref{fig:sadsample}, image emotion is related to complex visual features from high-level to low-level for both global and local views. Low-level visual features from the local view, such as color, shape, line and texture, were first used to classify image emotions \cite{kang2003affective,wang2008survey,aronoff2006we,hanjalic2006extracting}. Joshi \emph{et al.} \cite{joshi2011aesthetics} indicated that image emotion is highly related to image aesthetics for artistic works. Based on their study, mid-level features that represent image aesthetics, such as composition, visual balance and emphasis, are applied for image emotion classification \cite{machajdik2010affective,zhao2014exploring}. Machajdik and Hanbury suggested that image emotion can be significantly influenced by semantic content of the image \cite{machajdik2010affective}. They combined high-level image semantics from the global view with Itten's art theory on relevance of colors \cite{itten1962art} to recognize image emotion. However, most of the existing methods rely on hand-crafted features, which are manually designed based on common sense and observation of people. These methods can hardly take all important factors related to image emotion, i.e., image semantics, image aesthetics and low-level visual features, into consideration.

\begin{figure}[!htb]
    \begin{center}
    \includegraphics[width=1\linewidth]{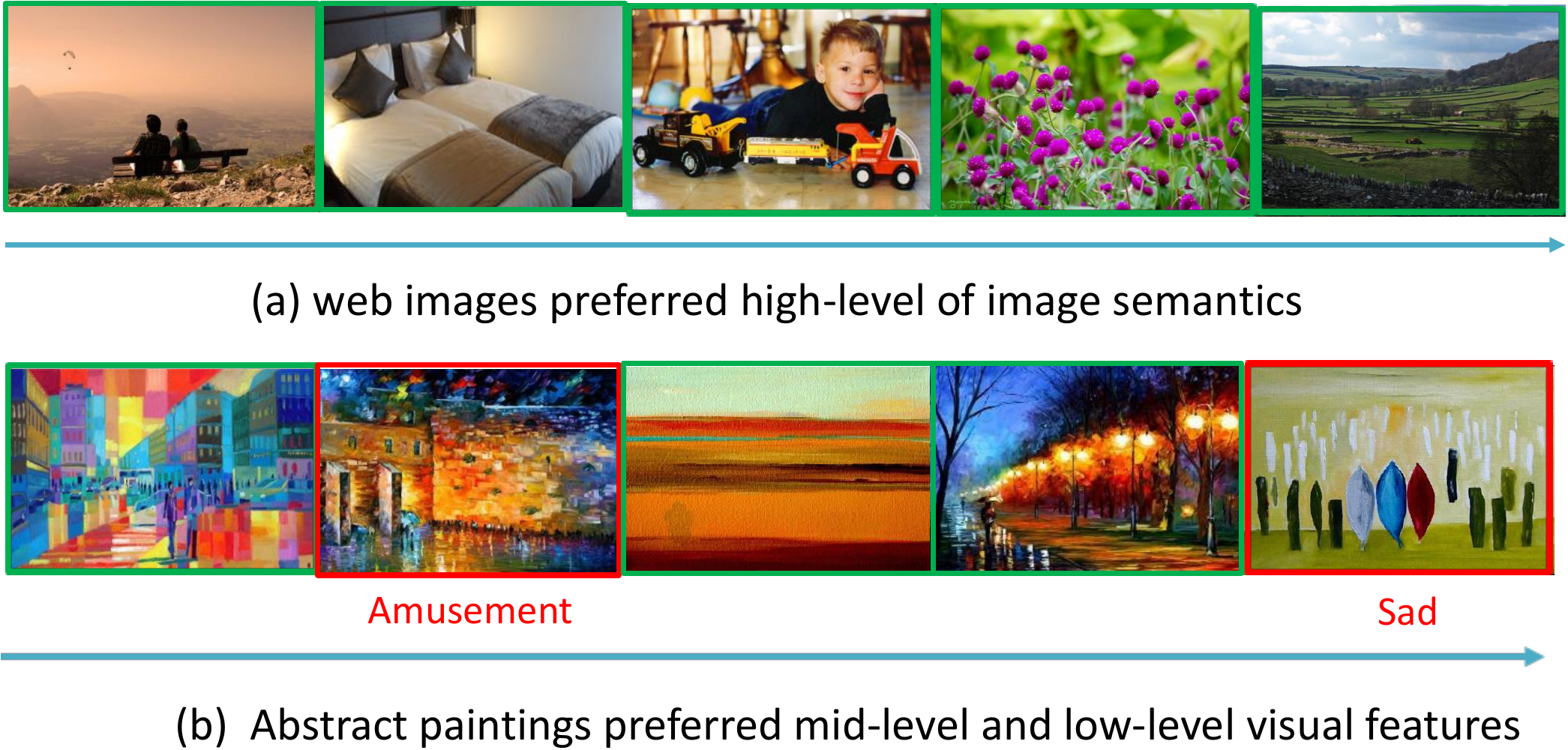}
    \end{center}
    \caption{Top 5 classification results for emotion category \emph{contentment} using AlexNet \cite{krizhevsky2012imagenet} on web images and abstract paintings. \emph{Green (Red)} box means correct (wrong) results, the correct label for wrong retrieve are provided. It is clear that AlexNet produces better matches for web images than abstract paintings. This means AlexNet deals high-level image semantics better than mid-level and low-level visual features.}
    \label{fig:topsample}
\end{figure}

Recently, with the rapid popularity of Convolutional Nerual Network (CNN), outstanding breakthroughs have been achieved in many visual recognition tasks, such as image classification \cite{krizhevsky2012imagenet}, image segmentation \cite{long2015fully}, object detection \cite{ren2015faster} and scene recognition \cite{zhou2014learning}. Instead of designing visual features manually, CNN provides an end-to-end feature learning framework, which can automatically learn deep representations of images from global view. Several researchers have also applied CNN to image emotion classification. However, as shown in Figure \ref{fig:topsample}, the currently used CNN methods, such as AlexNet \cite{krizhevsky2012imagenet}, for visual recognition cannot well deal with mid-level image aesthetics and low-level visual features from local view. In \cite{alameda2016recognizing}, the authors indicated that AlexNet is not effective enough to extract emotion information from abstract paintings, whose emotions are mainly conveyed by mid-level image aesthetics and low-level visual features.

What's more, the CNN-based methods usually rely on the large scale manually labeled training datasets like the ImageNet dataset \cite{deng2009imagenet}. People coming from different culture background may have very different emotion reactions to a particular image. Therefore, the emotional textual context associated with Internet images, e.g., titles, tags and descriptions, may not be reliable enough, and result the datasets collected from the Internet for emotion classification may contain noisy and inaccurate emotion labels. The emotion classification accuracy using existing methods, such as AlexNet, could be degraded when using these noisy labeled data as training data \cite{you2016building}.

\begin{figure*}[t]
    \begin{center}
    \includegraphics[width=1\linewidth]{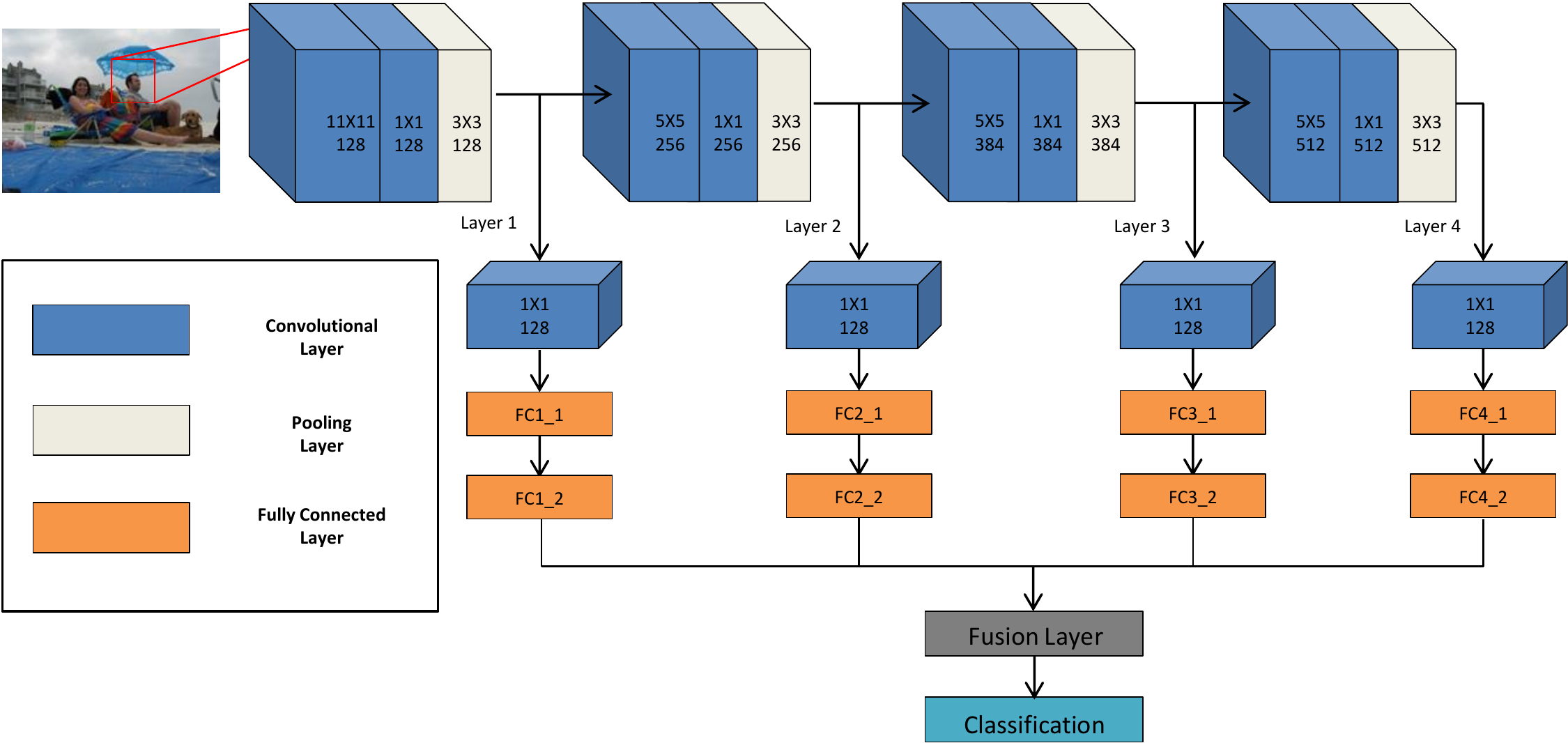}
    \end{center}
    \caption{Overview of our multi-level deep representation network (MldrNet). Different levels of deep representations related to high-level, mid-level and low-level visual features are extracted from different convolutional layer and fuse using fusion layer. The fusion representations are finally used for classification}
    \label{fig:Overview}
\end{figure*}

Considering the above mentioned two challenges, in this paper, we propose a new deep network (MldrNet) that learns multi-level deep representations from both global and local views for image emotion classification. Figure \ref{fig:Overview} shows an overview of the proposed MldrNet network. The traditional CNN method is designed for center-position object classification, which cannot effectively extract mid-level image aesthetics and low-level visual features from the local view. To perform end-to-end learning methods for different levels of deep representation from an entire image, we propose a CNN model with side branches to extract different levels of deep representations. Through a fusion layer, different levels of deep representations are integrated for classification. We notice that different fusion methods will largely effect the classification results when using noisy labeled data as training data. To demonstrate the effectiveness of our MldrNet and explore the impact of different fusion methods, we conduct extensive experiments on several publicly available datasets for different kinds of images, e.g. web images and abstract paintings.

The main contribution in our paper is that we propose a new CNN based method that combines different levels of deep representations, such as image semantics, image aesthetics and low-level visual features, from both global and local views for image emotion classification. Through combining different levels of deep representations, our method can effectively extract emotional information from images. Experimental results demonstrate that our method outperforms the state-of-the-art methods using deep features or hand-crafted features on both Internet images and abstract paintings.

The reminder of this paper is organized as follows. In section II, we will review the related works about visual emotion classification. Our proposed multi-level deep representation network for image emotion classification will be introduced in section III. Through various experiments on different kinds of datasets in section IV, we demonstrate that our method not only improves the emotion classification result compared to existing work, but also effectively deals with the noisy labeled data. Finally, we will conclude our work with future research directions in section V.

\section{Related Work}


Affective content analysis on multimedia data has been widely studied in recently years, including text \cite{hu2014semantic,cui2016sentiment}, audio \cite{shepstone2014using,poria2016fusing}, video \cite{hanjalic2005affective,soleymani2014corpus,yadati2014cavva} and image \cite{zhao2016continuous,sun2016sentiment}.

For visual emotion classification, existing research can be roughly divided into methods in dimensional emotion space (DES) \cite{xu2008,benini2011connotative,tarvainen2014content,zhao2016continuous} and methods in categorical emotion states (CES) \cite{machajdik2010affective,tang2012quantitative,zhao2014exploring,peng2015mixed}. DES models, which utilize 3-D valence-arousal-control emotion space, 3-D natural-temporal-energetic connotative space, 3-D activity-weight-heat emotion factors, and/or 2-D valence-arousal emotion space, provide predictable and flexible descriptions for emotions. In CES models, computational results are mapped directly to one of a few basic categories, such as \emph{anger}, \emph{excitement}, \emph{sadness}, \emph{etc}. Compared to DES models, CES models are straightforward for people to understand and label, thus have been widely applied in recent studies. To compare our result with existing work, we adopt CES model to classify emotions into eight categories predefined in a rigorous psychological study \cite{mikels2005emotional}.

The visual features used for image emotion classification are designed and extracted from different levels \cite{borth2013large,zhao2016predicting}. Yanulevskaya \emph{et al.} \cite{yanulevskaya2008} first proposed to categorize emotions of artworks based on low-level features, including Gabor features and Wiccest features. Solli and Lenz \cite{solli2009} introduced a color-based emotion-related image descriptor, which is derived from psychophysical experiments, to classify images. The impact of shape features for image emotion classification was discussed in \cite{lu2012shape}. In \cite{rao2016multi}, SIFT features extracted from both global view and local view were used for emotion prediction. Machajdik \emph{et al.} \cite{machajdik2010affective} defined a combination of rich hand-crafted mid-level features based on art and psychology theory, including \emph{composition}, \emph{color variance} and \emph{texture}. Zhao \emph{et al.} \cite{zhao2014exploring} introduced more robust and invariant mid-level visual features, which were designed according to art principles to capture information about image emotion. High-level adjective noun pairs related to object detection were introduced for visual sentiment analysis in recent years \cite{borth2013large,chen2014object}. Tkalcic \emph{et al.} \cite{tkalcic2013affective} indicated the emotional influence of facial emotion expressions for images and obtained the affective labels of images based on the high-level semantic content. However, these hand-crafted visual features have only been proven to be effective on several small datasets, whose images are selected from a few specific domains, e.g. abstract paintings and portrait photos. This limits the applications of image emotion classification on large-scale image datasets.

Considering the recent success from CNN-based approaches in many computer vision tasks, such as image classification \cite{krizhevsky2012imagenet}, image segmentation \cite{long2015fully}, object detection \cite{ren2015faster} and scene recognition \cite{zhou2014learning}, CNN based methods have also been employed in image emotion analysis. Peng \emph{et al.} \cite{peng2015mixed} first attempt to apply the CNN model in \cite{krizhevsky2012imagenet}. They finetune the pre-trained convolutional neural network on ImageNet \cite{deng2009imagenet} and show that CNN model outperforms previous methods rely on different levels of handcrafted features on Emotion6 dataset. You \emph{et al.} \cite{you2016building} combine CNN model in \cite{krizhevsky2012imagenet} with support vector machine (SVM) to detect image emotion on large-scale dataset of web images. These works usually borrow the popular CNN models that are used for image classification and object detection for image emotion classification. However, these widely used CNN models can not effectively classify the images whose emotion are mainly evoked by low-level and mid-level features, i.e. abstract paintings and art photos. Therefore, in this paper, we propose a new CNN model that can specifically deal with image emotion.

\section{The Proposed Method}

In this section, we introduce our method that learns multi-level deep representations (MldrNet) for image emotion classification. Consider image emotion is related to different levels of features, i.e., high-level image semantics, mid-level image aesthetics and low-level visual features, our method unifies different levels of deep representation within one CNN structure. In particular, we propose a fusion layer to support multi-level deep representations aggregation based on the characteristics of image emotion. Following the aforementioned discoveries, we divide the images into 8 emotion categories (positive emotion \emph{Amusement}, \emph{Awe}, \emph{Contentment}, \emph{Excitement} and negative emotion \emph{Anger}, \emph{Disgust}, \emph{Fear}, \emph{Sadness}) for visual emotion classification.

\subsection{Convolutional Neural Network}

Before introducing our MldrNet, we first review the CNN model that has been widely used for computer vision tasks \cite{krizhevsky2012imagenet}. Given one training sample $\left\{(x,y)\right\}$, where $x$ is the image and $y$ is the associated label, CNN extracts layer-wise representations of input images using convolutional layers and fully-connected layers. Followed by a softmax layer, the output of the last fully-connected layer can be transformed into a probability distribution $\textbf{p} \in \mathbb{R}^{m}$ for image emotions of $n$ categories. In this work, $n=8$ indicates eight emotion categories. The probability that the image belongs to a certain emotion category is defined blow:

\begin{equation}
    p_{i} = \frac{exp(h_{i})}{\sum_{i}exp(h_{i})}, i=1,...,n,
\end{equation}
where $h_{i}$ is the output from the last fully-connected layer. The loss of the predicting probability can also be measured by using cross entropy

\begin{equation}
    L = -\sum_{i}y_{i}\log(p_{i}),
\end{equation}
where $y=\{y_{i}|y_{i}\in \{0,1\},i=1,...,n,\sum_{i=1}^{n}p_{i}=1\}$ indicates the true emotion label of the image.

\begin{figure}
    \begin{center}
    \includegraphics[width=1\linewidth]{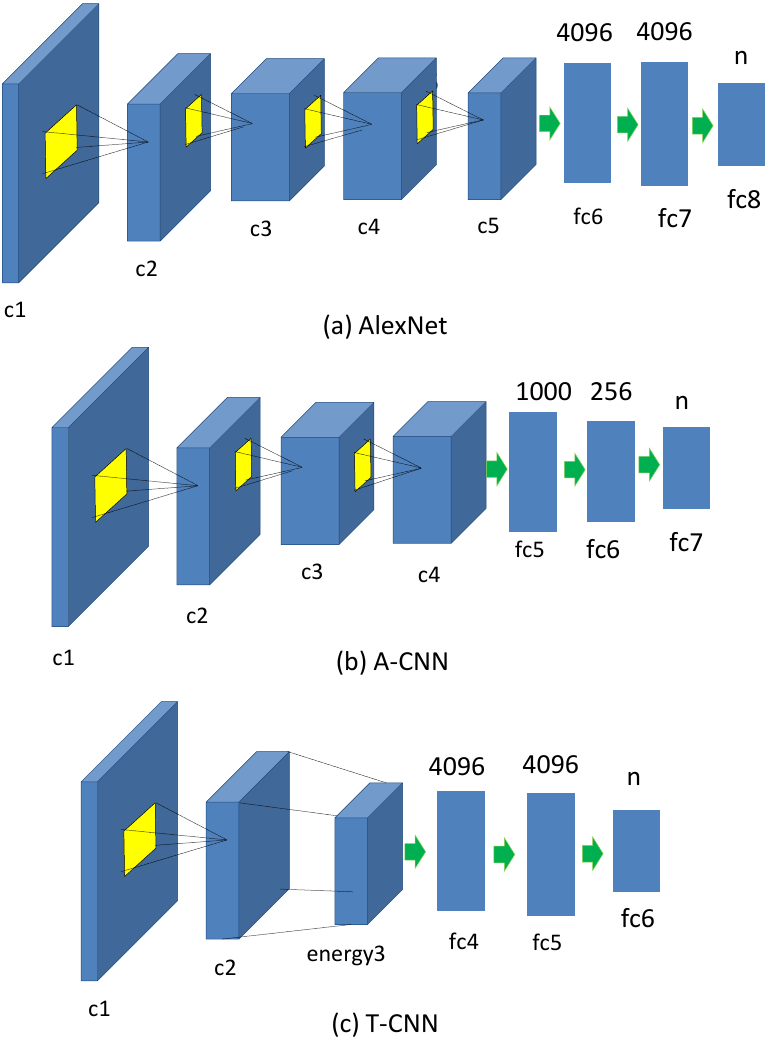}
    \end{center}
    \caption{The structures of different CNN models that deal with different levels of computer vision tasks.}
    \label{fig:structure}
\end{figure}

AlexNet is used for classify image on large scale dataset. It contains five convolutional layers followed by max-pooling layers, and three fully-connected layers, which contains 4,096, 4,096 and 8 neurons, respectively. The structure of AlexNet is shown in Fig \ref{fig:structure}(a). AlexNet is mainly trained for semantic-level image classification and tends to extract high-level deep representation about image semantics. It cannot effectively extract emotion information from abstract painting whose emotion is mainly convey by mid-level  image aesthetics and low-level visual features \cite{alameda2016recognizing}. As discussed in Section 1, AlexNet is likely not informative enough for image emotion classification.

\subsection{Analysis of different CNN models}

Emotion related image features can be roughly divided into low-level visual features, such as color, line and texture, mid-level image aesthetics, including composition and visual balance, and high-level image semantics. As CNN models contain a hierarchy of filters, the level of representations learned from CNN models are higher if one goes "deeper" in the hierarchy \cite{zeiler2014visualizing}. This means that if a CNN structure contains more convolutional layers, the level of feature extracted from the CNN structure is higher. Various CNN structures used in different computer vision tasks have also demonstrated this conclusion. To extract deep representations about mid-level image aesthetics and low-level visual features, different kinds of CNN models inspired by AlexNet, which contain less number of convolutional layers, are developed \cite{lu2014rapid,andrearczyk2016using}.

Image aesthetics has a close relationship with image emotion. To effectively deal with the mid-level image aesthetics, Aesthetics CNN(A-CNN) model has been developed \cite{lu2014rapid}
As shown in Figure \ref{fig:structure}(b), A-CNN consists of four convolutional layers and three fully-connected layers, which contains 1,000, 256 and 8 neurons, respectively. The first and second convolutional layers are followed by max-pooling layers. Even contains less convolutional layers compared to AlexNet, A-CNN has a better performance on image aesthetics analysis.

Texture has been proven as one of the important low-level visual features related to image emotion classification \cite{machajdik2010affective}. To extract deep representations about texture of images, an efficient CNN model, T-CNN, is designed for texture classification \cite{andrearczyk2016using}. As shown in Figure \ref{fig:structure}(c), T-CNN removes the last three conventional layers of AlexNet, and adds an energy layer (average-pooling layer with the kernel size as 27) behind the second convolutional layers. Following the energy layer, there are still three fully-connected layers, which contains 4,096, 4,096 and 8 neurons, respectively.

From the aforementioned CNN models we can find that the structures of different CNN models are similar, the main difference is the number of convolutional layers. This means we can share some parameters for different CNN models that can extract different levels of deep representations. Based on the observation, we unify different CNN models into one CNN structure, which can not only improve the classification accuracy of image emotion, but also gain a better parameter efficiency.

\subsection{Deep Network Learning Multi-level Deep representations}

\begin{figure}
    \begin{center}
    \includegraphics[width=1\linewidth]{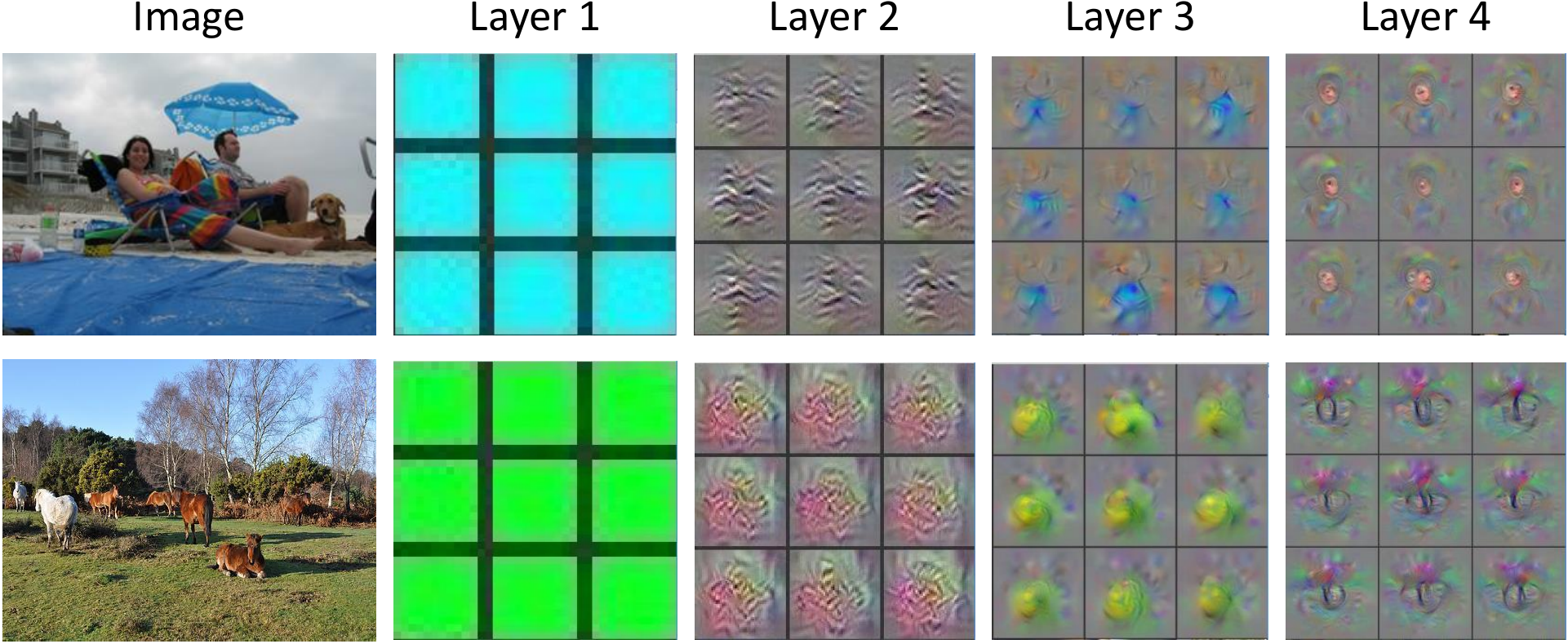}
    \end{center}
    \caption{Visualization of the weights of filter, which produce an activation map with the highest activation, in each convolutional layer.}
    \label{fig:RF}
\end{figure}

To effectively unify different levels of deep representations in one CNN model, we develop a multi-level deep representations network (MldrNet), which contains a main network and four side branches. Different levels of deep representation from both global view and local view can be extracted from different convolutional layers in our MldrNet. As shown in Figure \ref{fig:structure}, our MldrNet consists of 4 convolutional layers, whose filter size are $11\times11$, $5\times5$, $5\times5$ and $5\times5$, respectively. For each convolutional layer, it is followed by two fully connected layers. One problem for our MldrNet is that the dimension of the output of each convolutional layer is different. Inspired by the structure of GoogleNet \cite{szegedy2015going}, we insert a $1\times1$ convolutional layer with 128 filters between the pooling layer and fully connected layer for each layer of our MldrNet. The $1\times1$ convolutional layer unifies the output dimension of each layer and rectifies linear activation.

Compared to high-level image semantic information extracted from the highest layer of our MldrNet, deep representations extracted from the lower layers provide the additional information, such as low-level color and texture and mid-level composition and visual balance, which is related to image emotion. Existing research on image emotion analysis has demonstrated that, with the additional information related to low-level and mid-level image features, the performance of image emotion classification will be significantly improved \cite{zhao2014exploring}.

When designing the proposed MldrNet, two problems need to be considered. First, we should investigate the proper number of layers in our deep network. As we mentioned before, making the network deeper may not improve the image emotion classification results. If the number of layers is too high, the number of parameters will largely increased because each layer need to have its own weights, while the contribution of these layers to emotion classification may be very little. However, if the number of layers is too low, the extracted deep representations may not be able to well represent image emotion. To show the differences of various deep representations extracted from each layers in our network, we visualize the weights of the filter, which produces an activation map with the highest activation, in each convolutional layer in Figure \ref{fig:RF}. It is obvious that, the deep representations extracted from layer 1 and layer 2 are related to low-level features, while in layer 3, the deep representations focus on abstract concepts that reflect image aesthetics. In the highest layer, the deep representations mainly represent concrete objects in images, i.e. human face and horse. We also conduct experiments to investigate the impact of different number of layers in MldrNet for image emotion classification in Section \ref{sec:choice}

Second, the role of deep representations extracted from different layer of MldrNet for evoking emotions may vary for different kinds of images. To effectively combine the different levels of deep representations, the fuse function need to be carefully chosen. We introduce the most commonly used fusion function in our MldrNet, including concatenation, $min(\cdot)$, $max(\cdot)$ and $mean(\cdot)$. Detailed discussion of fusion layer will be shown in Section \ref{sec:fusion}.

\subsection{Fusion Layer} \label{sec:fusion}

Fusion layer is the core component of our multi-level deep representations network, which is comprised of a collection of fusion functions. As some image information will be disregarded when passing a convolutional layer, some existing models, i.e. ResNet \cite{szegedy2015going} and DenseNet \cite{he2016deep}, combines information from different convolutional layers to improve the performance. However, they just simply concatenate multi-level features through skip-connection, which means information extracted from different convolutional layers has equal weights. While in image emotion analysis, different level of features may have different impact on evoking emotions. To choose a suitable fusion function for image emotion classification result, we use different fusion functions in fusion layer to fuse multi-level deep representations.

We define the deep representation we extract from the $i$\emph{th} layer is $h_{i}$ and the fusion function is $f(x)$. Then the representation of the entire image can be aggregated by using the representation of each layer

\begin{equation}
    \hat{h}=f(\hat{h_{1}},\hat{h_{2}},...,\hat{h_{i}}).
\end{equation}

The distribution of emotion categories of the image and the loss $L$ can be computed as

\begin{equation}
    p_{i} = \frac{exp(\hat{h})}{\sum_{i}exp(\hat{h})} and  L = -\sum_{i}y_{i}\log(p_{i}),
\end{equation}

In our experiments, we have the fusion function $f(x)={min, max, mean}$. We can easily find out that the function $mean(\cdot)$ assigns the same weight to deep representations extracted from each convolutional layer, while the function $max(\cdot)$ and $min(\cdot)$ would encourage the model to increase the weight of one out of all layers of deep representations. The choice of the fusion function is of critical importance in our method. The comparison results of utilizing different fusion functions are shown in Section IV.

\section{Experiments}

In this section, we evaluate the performance of our MldrNet on different datasets. The recently published large scale dataset for emotion recognition \cite{you2016building} and three popular used small datasets: IAPS-Subset \cite{mikels2005emotional}, ArtPhoto and Abstract \cite{machajdik2010affective} are used to evaluate the classification results over 8 emotion categories. The MART dataset \cite{yanulevskaya2012eye} is used to evaluate the classification result on abstract paintings over 2 emotion categories (positive and negative).

\subsection{Experimental Settings}

\subsubsection{Implementation Details}

We implement our model by using the pyTorch framework on two Nvidia GTX1080. The detailed parameters of our model is presented in Fig \ref{fig:Overview} and the input images are cropped as $375\times375$ from center and corners. The batch size is set to 64. We optimize our model using stochastic gradient descent(SGD). The initial learning rate is empirically set as 0.001, the momentum is 0.9 , and weight decay is 0.0005. The parameters in these optimizers are initialized by using the default setting.

\subsubsection{Datasets}

\textbf{Large Scale Dataset For Emotion classification:} This dataset is newly published in \cite{you2016building} to evaluate the classification result over 8 different emotion categories (positive emotions \emph{Amusement}, \emph{Awe}, \emph{Contentment}, \emph{Excitement} and negative emotions \emph{Anger}, \emph{Disgust}, \emph{Fear}, \emph{Sad}). To collect this dataset, 90,000 noisy labeled images are first downloaded from Instagram and Flickr by using the names of emotion categories as the key words for searching. Then, the downloaded images were submitted to Amazon Mechanical Turk (AMT) for further labeling. Finally, 23,308 well labeled images were collected for emotion recognition \footnote {We have 88,298 noisy labeled images and 23,164 manually labeled images as some images no longer exists in the Internet.}.

\textbf{Small Scale Datasets For Emotion Classification}: Three small datasets that are widely used in previous works for image emotion classification are introduced below.

(1)\textbf{IAPS-Subset}: The \emph{International Affective Picture System} (IAPS) is a standard stimulus image set which has been widely used in affective image classification. IAPS consists of 1,182 documentary-style natural color images depicting complex scenes, such as portraits, puppies, babies, animals, landscapes and others \cite{lang2008international}. Among all IAPS images, Mikels \emph{et al.} \cite{mikels2005emotional} selected 395 images and mapped arousal and valence values of these images to the above mentioned eight discrete emotion categories.

(2)\textbf{ArtPhoto}: In the ArtPhoto dataset, 806 photos are selected from some art sharing sites by using the names of emotion categories as the search terms \cite{machajdik2010affective}. The artists, who take the photos and upload them to the websites, determine emotion categories of the photos. The artists try to evoke a certain emotion for the viewers of the photo through the conscious manipulation of the emotional objects, lighting, colors, etc. In this dataset, each image is assigned to one of the eight aforementioned emotion categories.

(3)\textbf{Abstract}: This dataset consists of 228 abstract paintings. Unlike the images in the IAPS-Subset and ArtPhoto dataset, the images in the Abstract dataset represent the emotions through overall color and texture, instead of some emotional objects \cite{machajdik2010affective}. In this dataset, each painting was voted by 14 different people to decide its emotion category. The emotion category with the most votes was selected as the emotion category of that image.

\textbf{MART}: The MART dataset is a collection of 500 abstract paintings from the Museum of Modern and Contemporary Art of Trento and Rovereto. These artworks were realized since the beginning of the 20 century until 2008 by professional artists, who have theoretical studies on art elements, such as colors, lines, shapes and textures, and reflect the results of studies on their paintings. Using the relative score method in \cite{sartori2015s}, the abstract paintings are labeled as positive or negative according to the emotion type evoked by them.

\subsubsection{Compared Methods}

To demonstrate the effectiveness of our MldrNet, we compared our method with state-of-the-art image emotion classification methods and the most popular CNN models:

\textbf{Machajdik}\cite{machajdik2010affective}: using the low-level visual features and mid-level features inspired by psychology and art theory which are specific to art works.

\textbf{Zhao}\cite{zhao2014exploring}: using principles-of-art-based emotion features, which are the unified combination of representation features derived from different principles, including \emph{balance}, \emph{emphasis}, \emph{harmony}, \emph{variety}, \emph{gradation}, and \emph{movement} and its influence on image emotions.

\textbf{Rao}\cite{rao2016multi}: using different visual features extracted from multi-scale image patches.

\textbf{AlexNet+SVM}\cite{you2016building}: using AlexNet to extract emotion related deep features and classify them through SVM.

\textbf{AlexNet}\cite{krizhevsky2012imagenet}: pre-trained based on ImageNet and fine-tuned using the large scale dataset for emotion classification.

\textbf{VGGNet-19}\cite{Simonyan14c}: pre-trained based on ImageNet and fine-tuned using the large scale dataset for emotion classification.

\textbf{ResNet-101}\cite{he2016deep}: pre-trained based on ImageNet and fine-tuned using the large scale dataset for emotion classification.

To fully quantify the role of different fusion functions of our model and detect the suitable architecture of our model, different variants of our model are compared:

\textbf{MldrNet-concat}: simply concatenate deep representations extracted from each layer in the fusion layer.

\textbf{MldrNet-max}: using $max$ as fusion function in the fusion layer.

\textbf{MldrNet-min}: using $min$ as fusion function in the fusion layer.

\textbf{MldrNet-mean} using $mean$ as fusion function in the fusion layer.

\subsection{Emotion Classification on Large Scale and Noisy Labeled Dataset}\label{sec:choice}

The well labeled 23,164 images are randomly split into the training set (80\%, 18,532 images), the testing set (15\%, 3,474 images) and the validaton set (5\%, 1,158 images). Meanwhile, to demonstrate the effectiveness of our approach on noisy labeled dataset, we create a noisy labeled dataset for training by combining the images, which have been submitted to AMT for labeling but labeled from different emotion categories, with the training set of well labeled images. The noisy labeled dataset contains 83,664 images for training. We called the well labeled dataset as \emph{well} dataset and noisy labeled dataset as \emph{noisy} dataset. The \emph{well} dataset and \emph{noisy} dataset are used for training models. The testing dataset is used to test our models.

\subsubsection{Choice of Number of Layers in MldrNet}

Our MldrNet model can utilize multi-level deep representations to classify image emotion by increasing or decreasing the number of convolutional layers. Choosing a proper number of convolutional layers in MldrNet needs to be explored in order to achieve the best emotion classification performance. We conduct experiments using MldrNet models with different number of convolutional layer.

\begin{table}[!htb]
\begin{center}
\begin{tabular}{|c|c|}
\hline
Model & Accuracy   \\ \hline
MldrNet-2 layer & 52.12\% \\ \hline
MldrNet-3 layer & 58.34\% \\ \hline
MldrNet-4 layer & 67.24\% \\ \hline
MldrNet-5 layer & 67.55\% \\ \hline
MldrNet-6 layer & 67.68\% \\ \hline
\end{tabular}
\end{center}
\caption{Emotion classification accuracy for MldrNet Models of different number of convolutional layer.}
\label{table:levelnumber}
\end{table}

As shown in Table \ref{table:levelnumber}, changing the number of convolutional layers in our MldrNet model will affect the classification accuracy. The models with fewer layers perform worse than the models with more than 4 layers. The main reason may be that the models with fewer layers lack the information related to high-level image features. What's more, the models with more than 4 layers cannot significantly improve the emotion classification accuracy, which indicates the contribution of these layers may be very little. Meanwhile, with more convolutional layers, the number of parameters needed to be computed is increased, therefore, the time for training these models are largely increased. Due to the above considerations, MldrNet with 4 layers is the best model the following experiments.

\subsubsection{Choice of Fusion Function}

Another important component in our MldrNet model is the fusion layer. As discussed previously, fusion function will effect the emotion classification accuracy. We also find that fusion function plays an important role for dealing with different training datasets.

\begin{table}[!htb]
\begin{center}
\begin{tabular}{|c|c|c|}
\hline
\multirow{2}{*}{Model}& \multicolumn{2}{c|}{Accuracy}   \\ \cline{2-3}
&\emph{well} dataset& \emph{noisy} dataset \\ \hline
MldrNet-concat & 66.92\% & 55.34\% \\ \hline
MldrNet-max    & 64.79\% & 53.68\% \\ \hline
MldrNet-min    & 62.44\% & 49.32\% \\ \hline
MldrNet-mean   & \textbf{67.24\%} & \textbf{59.85\%} \\ \hline
\end{tabular}
\end{center}
\caption{Emotion classification accuracy for MldrNet Models of different fusion function training on both \emph{well} dataset and \emph{noisy} dataset.}
\label{table:function}
\end{table}

In Table \ref{table:function}, we present the result of our MldrNet with a variety of fusion functions using both \emph{well} dataset and \emph{noisy} dataset for training. We notice that, compare to the MldrNet model of using  $max(\cdot)$ and $min(\cdot)$ as fusion function, the performances of MldrNet model of using $mean(\cdot)$ and $concat(\cdot)$ as fusion functions are better. Especially for fusion function $mean(\cdot)$, our MldrNet achieves the best performance when using different training dataset. Unlike $max(\cdot)$ and $min(\cdot)$ , when using the $mean(\cdot)$ and $concat(\cdot)$ as fusion function, the model can keep more emotional information extracted from each convolutional layers. Using $mean(\cdot)$ as fusion function in our model can better fuse the emotional information for image emotion classification.

\subsubsection{Comparison of Different Methods}

To investigate the effectiveness of our MldrNet model, we compared our model with different image emotion classification methods, including the state-of-the-art method using hand-crafted features and popular deep learning models. All methods use \emph{well} dataset as training dataset. The results are shown in Table \ref{table:overall}.

\begin{table}[!htb]
\begin{center}
\begin{tabular}{|c|c|}
\hline
Methods     & Accuracy \\ \hline
Zhao        & 46.52\%  \\ \hline
Rao         & 51.67\%  \\ \hline
AlexNet-SVM & 57.89\%  \\ \hline
AlexNet     & 58.61\%  \\ \hline
VGGNet-19   & 59.32\%  \\ \hline
ResNet-101  & 60.82\%  \\ \hline
MldrNet     & 67.24\%  \\ \hline
MldrNet+Bi-GRU& 73.03\% \\ \hline
\end{tabular}
\end{center}
\caption{Emotion classification accuracy for different methods on the large scale dataset for image emotion classification.}
\label{table:overall}
\end{table}

From Tabel \ref{table:overall}, we have the following observations. First of all, methods using deep representations outperforms the methods using hand-crafted features. These hand-crafted features are designed based on several small scale datasets composed by images from specific domains, which cannot comprehensively describe image emotion compared to deep representations. Then, for methods using deep representations, we can find that, compared to AlexNet, even though containing more convolutional layers and providing higher deep representations in VGGNet-19 and ResNet-101, the performances are just slightly improved. Only containing 4 convolutional layers, our MldrNet model considering both mid-level and low-level deep representations significantly improves the emotion classification accuracy, compared with other 'deeper' CNN models. Finally, when using noisy dataset for training, our MldrNet model can still achieve competitive emotion classification accuracy. This means our method can utilize the images which are directly collected for the Internet, which makes our method can be applied for many applications, such as, recommending system, social network and personalize advertising.

\begin{figure}
    \begin{center}
    \includegraphics[width=1\linewidth]{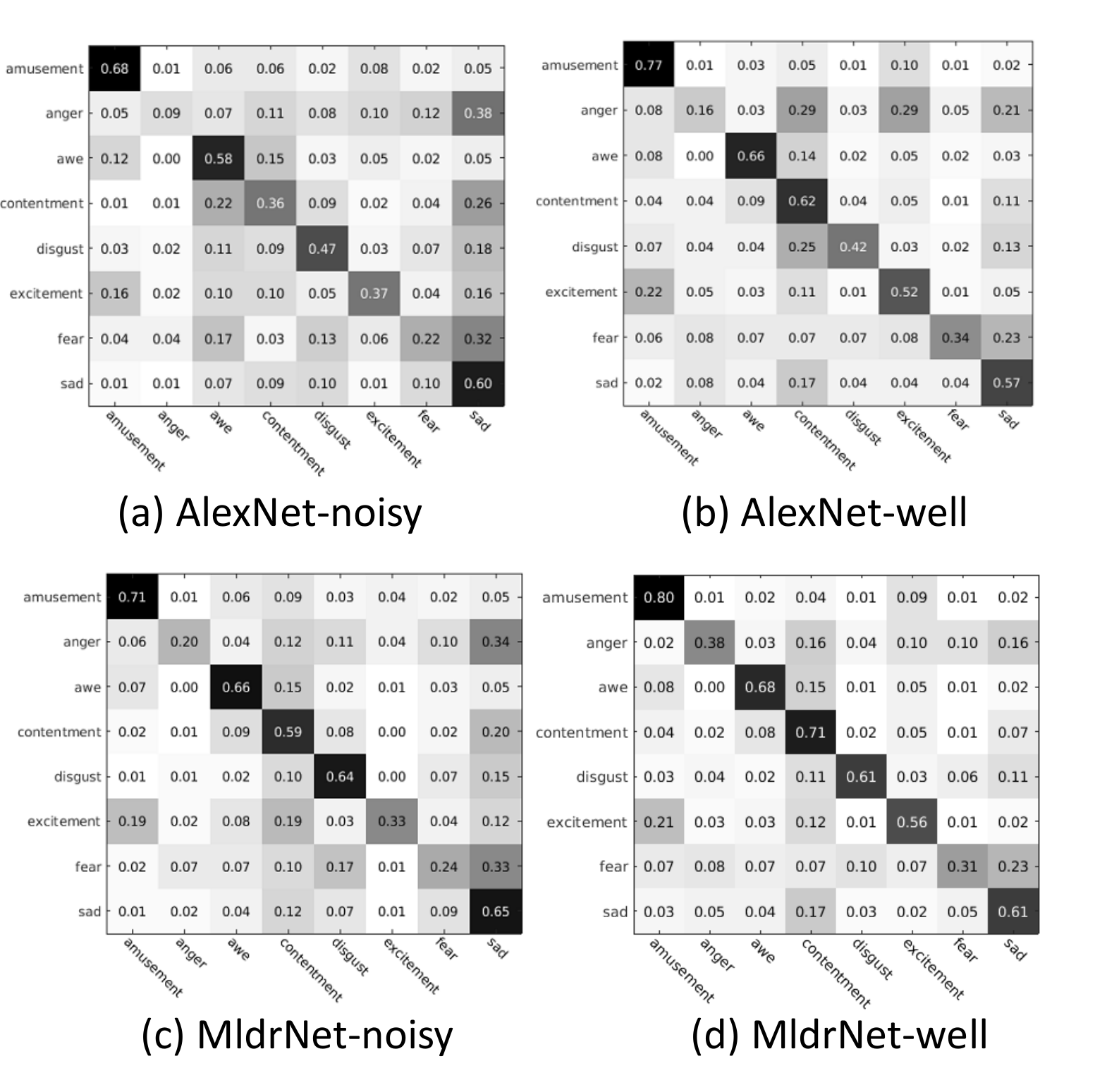}
    \end{center}
    \caption{Confusion matrices for AlexNet and our MldrNet when using the \emph{well} dataset and the \emph{noisy} dataset as training dataset.}
    \label{fig:cm}
\end{figure}

To further compared our methods with AlexNet, we report the confusion matrix of the two methods on the testing dataset. Considering the significant performance improvements by using deep representations compared to hand-crafted features, we only show the confusion matrices of our MldrNet and AlexNet using the \emph{well} dataset as the training dataset (MldrNet-well and AlexNet-well) and the \emph{noisy} dataset as the training dataset (MldrNet-noisy and AlexNet-noisy). As shown in Figure \ref{fig:cm}, the performances of AlexNet using both \emph{well} and \emph{noisy} as training dataset in most emotional categories are lower than our MldrNet. AlexNet tend to confuse some emotions, such as \emph{fear} and \emph{sad}. This indicates that image emotions cannot be clearly analyzed only relying on high-level image semantics. What's more, compared to Alexnet, our MldrNet shows a more robust emotion classification result when using different training dataset. This means our MldrNet can effectively extract emotional information even using training dataset that contains false label.

\begin{figure*}[!htb]
    \begin{center}
    \includegraphics[width=1\linewidth]{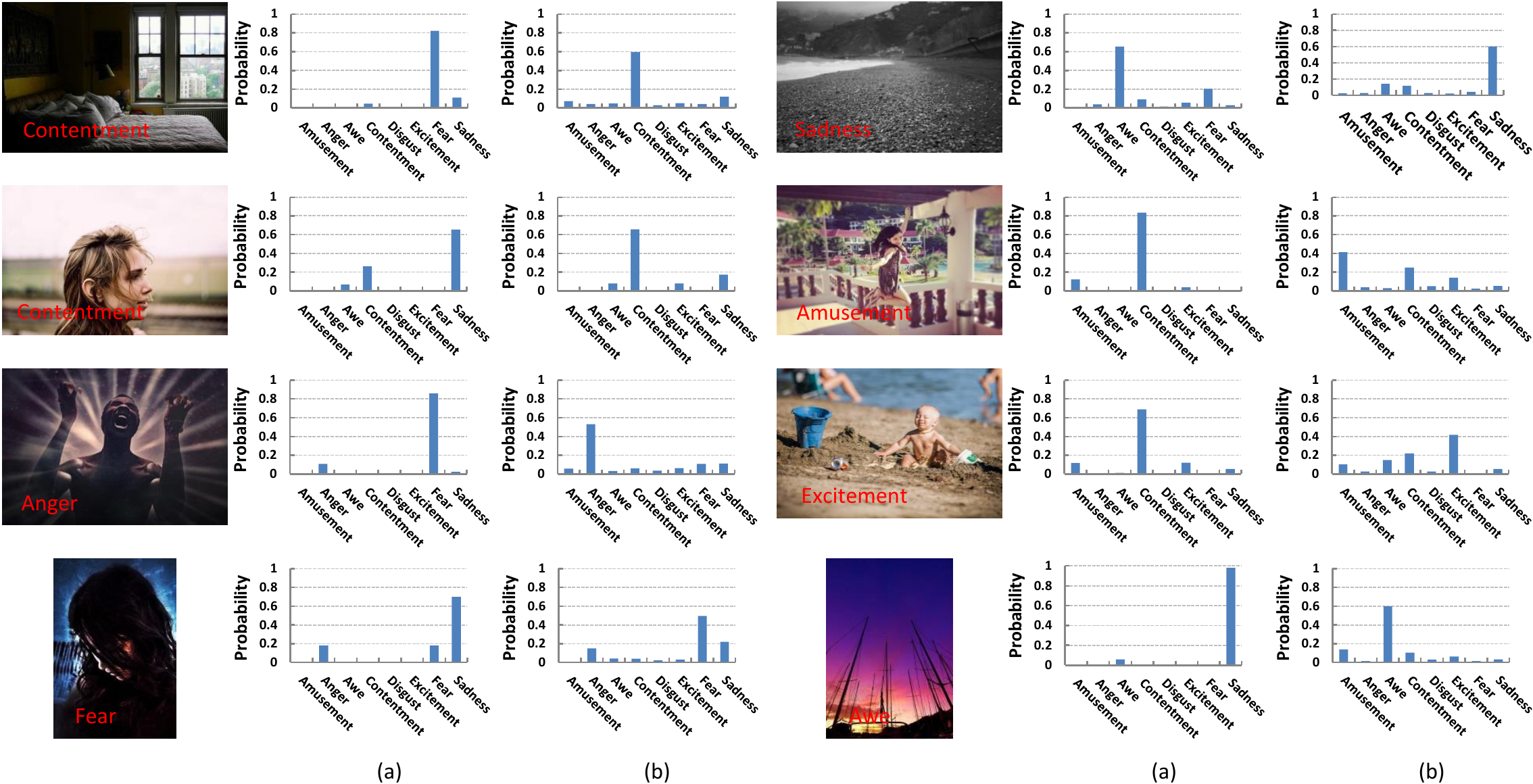}
    \end{center}
    \caption{Sample images correctly classified by our MldrNet but misclassified by AlexNet. The column (a) shows the emotion distribution predicted by AlexNet and the column (b) shows the emotion distribution predicted by our MldrNet. The red label on each image indicates the ground-truth emotion category.}
    \label{fig:sample}
\end{figure*}

We also visualize a couple of sample images that are correctly classified by our MldrNet but incorrectly classified by AlexNet to qualitatively analyze the influence of mid-level and low-level deep representations for image emotion classification. As shown in Figure \ref{fig:sample}, the emotions of the images misclassified by AlexNet are mainly convey by mid-level and low-level visual features, such as color, texture and image aesthetics. Combining the emotion information related to mid-level and low-level deep representations can significantly improve the emotion classification accuracy.


\subsection{Emotion Classification on small Scale Datasets}

We have introduced several image emotion analysis methods using hand-crafted features. To better evaluate the effectiveness of MldrNet, we compare our method with state-of-the-art methods based on hand-crafted features and Alexnet for each emotion categories.

\begin{figure}[!htb]
    \begin{center}
    \includegraphics[width=1\linewidth]{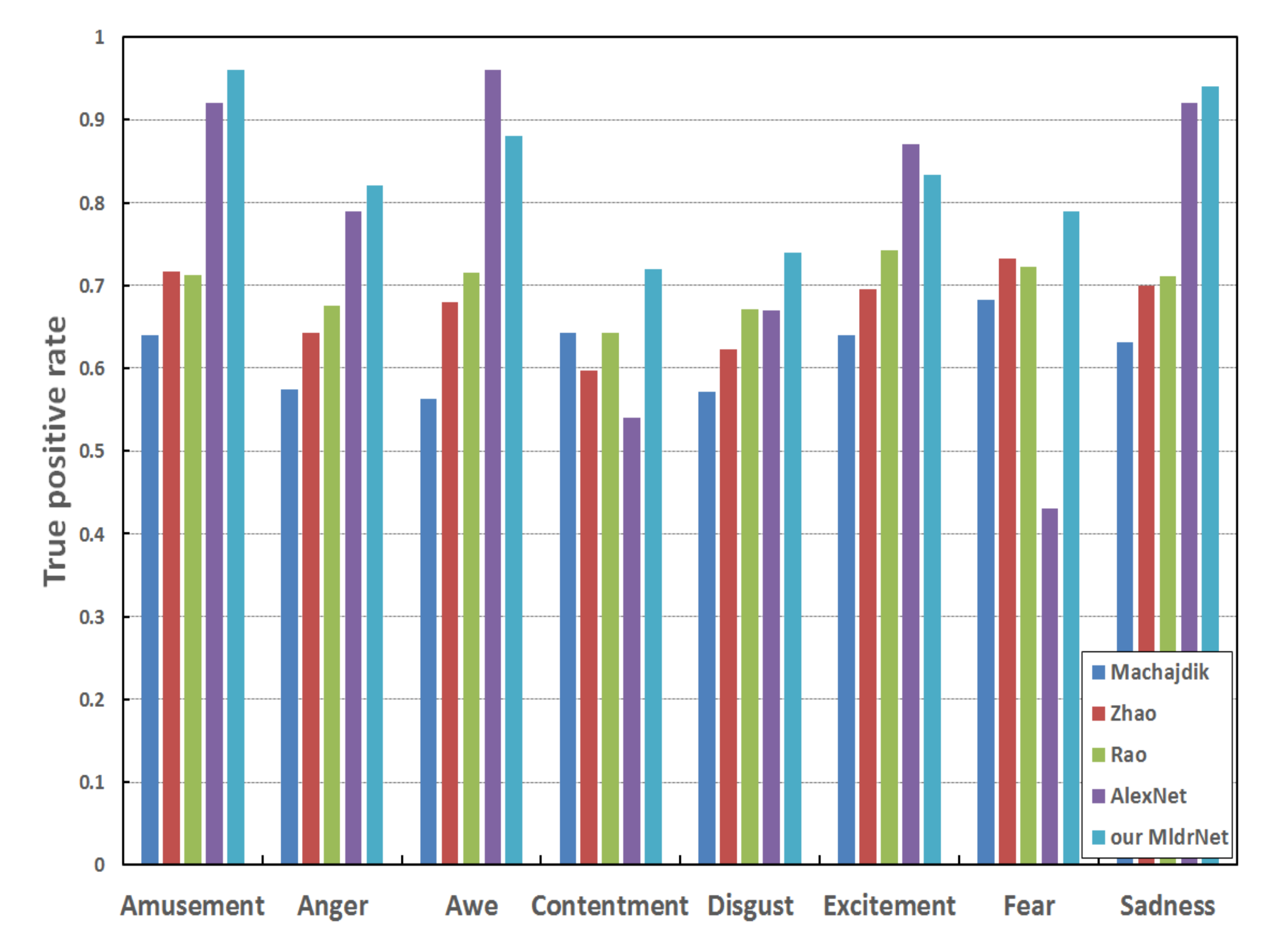}
    \end{center}
    \caption{Performance evaluation for each emotion categories  on the ArtPhoto dataset.}
    \label{fig:artphoto}
\end{figure}

\begin{figure}[!htb]
    \begin{center}
    \includegraphics[width=1\linewidth]{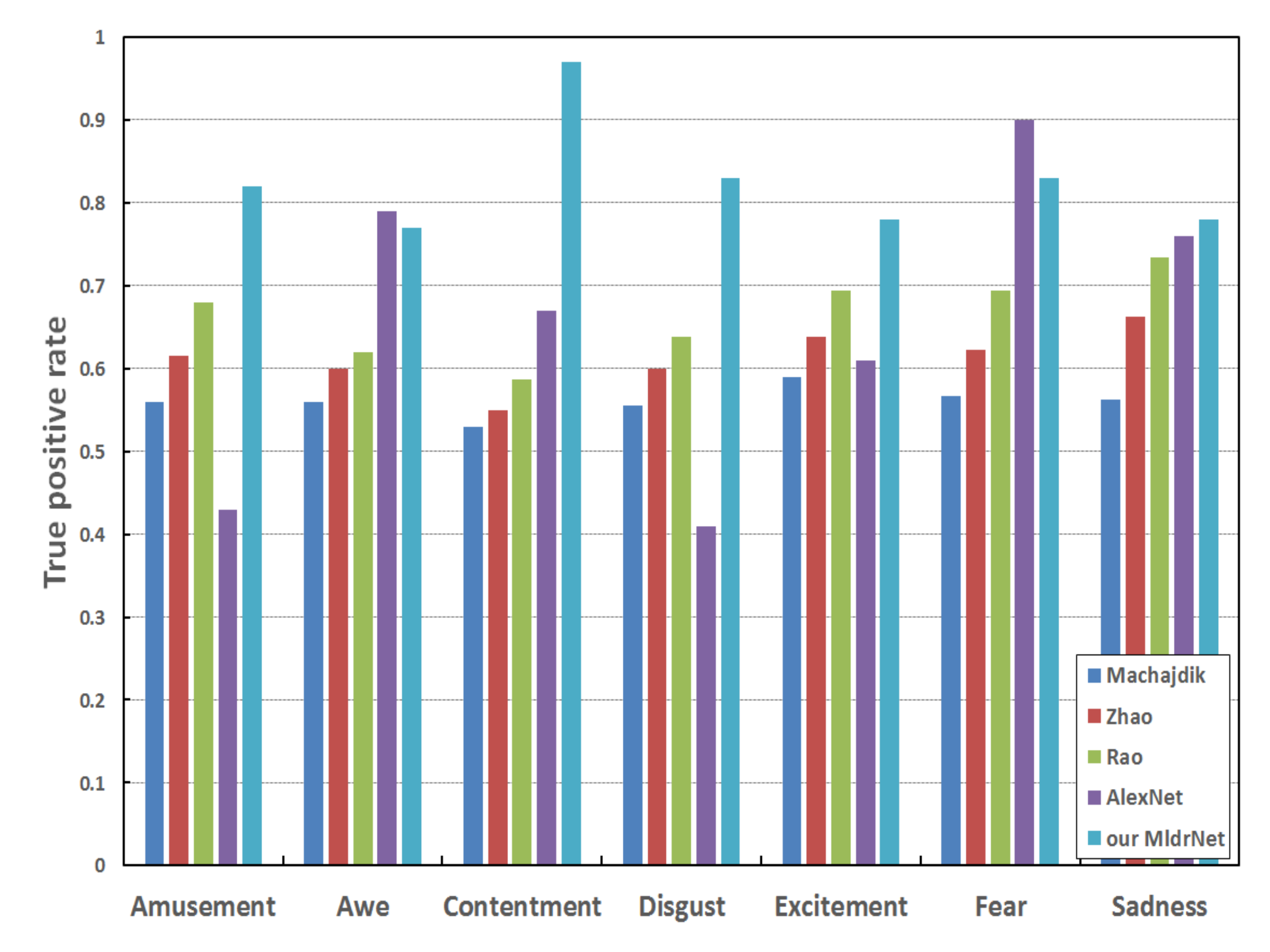}
    \end{center}
    \caption{Performance evaluation for each emotion categories  on the Abstract dataset.}
    \label{fig:abstract}
\end{figure}

\begin{figure}[!htb]
    \begin{center}
    \includegraphics[width=1\linewidth]{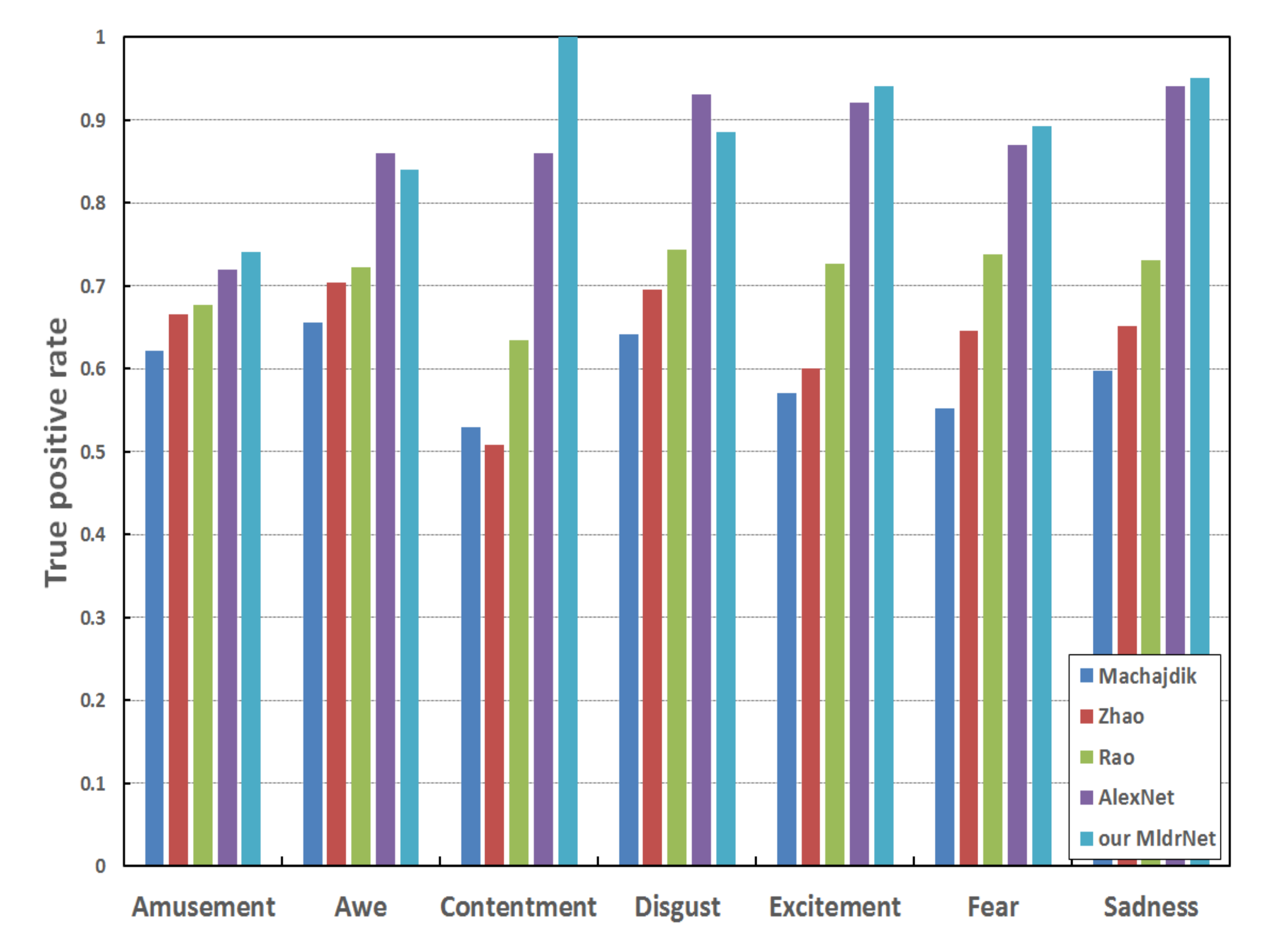}
    \end{center}
    \caption{Performance evaluation for each emotion categories on the IAPS-Subset.}
    \label{fig:iaps}
\end{figure}

We follow the same experimental settings described in \cite{machajdik2010affective}. Due to the imbalanced and limited number of images per emotion category, we employ the ``one against all'' strategy to train the classifier. The image samples from each category are randomly split into five batches and 5-fold cross validation strategy is used to evaluate the different methods. We use the images to train the last fully connected layer in our MldrNet and AlexNet. Also, the \emph{true positive rate per class} suggested in \cite{machajdik2010affective} is calculated to compare the results. Note that in IAPS-Subset and Abstract dataset, only eight and three images are contained in the emotion category \emph{anger}, so we are unable to perform 5-fold cross validation for this category. Therefore, the \emph{true positive rate per class} of emotion category \emph{anger} in these two datasets is not reported.

The emotion classification accuracies for each emotion categories are reported in Figure \ref{fig:artphoto},\ref{fig:abstract} and \ref{fig:iaps}, respectively. For most of emotion categories, deep learning methods significantly outperform the state-of-the-art hand-crafted methods. However, the performances of AlexNet in Abstract and ArtPhoto dataset are relatively low, this may because emotions of images in these two datasets are mainly conveyed by mid-level and low-level visual features. In contrast, MldrNet achieves the best performance in almost all emotion categories for the three dataset, which shows a robust result.

\subsection{Emotion Classification on Abstract Paintings}

To further evaluate the benefits of MldrNet. We also test our MldrNet on the MART dataset, which consists of abstract paintings. Followed the experimental approach in \cite{alameda2016recognizing}, we employ 10-fold cross validation to compare our MldrNet model with other six baseline methods on the MART dataset. The baseline methods are: kernel transductive SVM (TSVM \cite{joachims1999transductive}), linear matrix completion (LMC \cite{chen2015matrix}),  Lasso and Group Lasso both proposed in \cite{sartori2015s}, non-linear matrix completion (NLMC \cite{alameda2016recognizing}) and AlexNet \cite{krizhevsky2012imagenet}. The results shown in Table \ref{table:painting} demonstrate that our MldrNet can effectively extract emotion information from abstract paintings when compared with all other methods. Compared to traditional CNN models, MldrNet Model is especially good at dealing with the image emotion related to low-level and mid-level visual features, i.e. color, texture and image aesthetics.

\begin{table}
\begin{center}
\begin{tabular}{|c|c|}
\hline
Model & Accuracy  \\ \hline
TSVM  & 69.2\%  \\ \hline
LMC   & 71.8\% \\ \hline
Lasso  &  68.2\% \\ \hline
Group Lasso  & 70.5\% \\ \hline
NLMC  &72.8\% \\ \hline
AlexNet  & 69.8\% \\ \hline
MldrNet & 76.4\% \\ \hline
\end{tabular}
\end{center}
\caption{Emotion classification accuracy of different methods on the MART dataset.}
\label{table:painting}
\end{table}

\section{Conclusion}

In this paper, we have proposed a new network that learns multi-level deep representations for image emotion classification. We have demonstrated that image emotion is not only affected by high-level image semantics, but also related mid-level and low-level visual features, such as color, texture and image aesthetics. Our MldrNet successfully combine the deep representations extracted from different layer of deep convolutional network for image emotion classification. In our experiments, MldrNet achieves consistent improvement in image emotion classification accuracy with fewer convolutional layers compared to popular CNN models for different kind of image emotion datasets. Moreover, MldNet shows a more robust results when using different training dataset, especially the \emph{noisy} dataset directly collected from the Internet. This will decrease the demand for reliable training data, which will help us to utilize huge amount of images.

Compared to linear deep convolutional neural network models, we believe MldrNet model combining with deep representations extracted from different convolutional layers are better at dealing with abstract-level computer vision tasks, i.e. image emotion classification, image aesthetics analysis and photo quality assessment. In the future, we will extend the application of MldrNet in other abstract-level computer vision tasks. We also plan to explore the rules of image emotion for different vision tasks.


%




\ifCLASSOPTIONcaptionsoff
  \newpage
\fi


\begin{thebibliography}{10}

\bibitem{lang1979bio}
P.~J. Lang, ``A bio-informational theory of emotional imagery,'' {\em
  Psychophysiology}, vol.~16, no.~6, pp.~495--512, 1979.

\bibitem{joshi2011aesthetics}
D.~Joshi, R.~Datta, E.~Fedorovskaya, Q.-T. Luong, J.~Z. Wang, J.~Li, and
  J.~Luo, ``Aesthetics and emotions in images,'' {\em IEEE Signal Processing
  Magazine}, vol.~28, no.~5, pp.~94--115, 2011.

\bibitem{machajdik2010affective}
J.~Machajdik and A.~Hanbury, ``Affective image classification using features
  inspired by psychology and art theory,'' in {\em ACM MM}, pp.~83--92, 2010.

\bibitem{zhao2014exploring}
S.~Zhao, Y.~Gao, X.~Jiang, H.~Yao, T.-S. Chua, and X.~Sun, ``Exploring
  principles-of-art features for image emotion recognition,'' in {\em ACM MM},
  2014.

\bibitem{rao2016multi}
T.~Rao, M.~Xu, H.~Liu, J.~Wang, and I.~Burnett, ``Multi-scale blocks based
  image emotion classification using multiple instance learning,'' in {\em
  ICIP}, 2016.

\bibitem{zhao2016continuous}
S.~Zhao, H.~Yao, Y.~Gao, R.~Ji, and G.~Ding, ``Continuous probability
  distribution prediction of image emotions via multi-task shared sparse
  regression,'' {\em IEEE Transactions on Multimedia}, vol.~19, no.~3,
  pp.~632--645, 2017.

\bibitem{wei2004image}
W.~Wei-ning, Y.~Ying-lin, and Z.~Jian-chao, ``Image emotional classification:
  static vs. dynamic,'' in {\em SMC}, 2004.

\bibitem{kang2003affective}
H.-B. Kang, ``Affective content detection using hmms,'' in {\em ACM MM}, 2003.

\bibitem{wang2008survey}
W.~Wang and Q.~He, ``A survey on emotional semantic image retrieval.,'' in {\em
  ICIP}, 2008.

\bibitem{aronoff2006we}
J.~Aronoff, ``How we recognize angry and happy emotion in people, places, and
  things,'' {\em Cross-cultural research}, vol.~40, no.~1, pp.~83--105, 2006.

\bibitem{hanjalic2006extracting}
A.~Hanjalic, ``Extracting moods from pictures and sounds: Towards truly
  personalized tv,'' {\em IEEE Signal Processing Magazine}, vol.~23, no.~2,
  pp.~90--100, 2006.

\bibitem{itten1962art}
J.~Itten and E.~Van~Haagen, {\em The Art of Color; the Subjective Experience
  and Objective Rationale of Colour}.
\newblock Reinhold, 1962.

\bibitem{krizhevsky2012imagenet}
A.~Krizhevsky, I.~Sutskever, and G.~E. Hinton, ``Imagenet classification with
  deep convolutional neural networks,'' in {\em NIPS}, 2012.

\bibitem{long2015fully}
J.~Long, E.~Shelhamer, and T.~Darrell, ``Fully convolutional networks for
  semantic segmentation,'' in {\em CVPR}, 2015.

\bibitem{ren2015faster}
S.~Ren, K.~He, R.~Girshick, and J.~Sun, ``Faster r-cnn: Towards real-time
  object detection with region proposal networks,'' in {\em NIPS}, 2015.

\bibitem{zhou2014learning}
B.~Zhou, A.~Lapedriza, J.~Xiao, A.~Torralba, and A.~Oliva, ``Learning deep
  features for scene recognition using places database,'' in {\em NIPS}, 2014.

\bibitem{alameda2016recognizing}
X.~Alameda-Pineda, E.~Ricci, Y.~Yan, and N.~Sebe, ``Recognizing emotions from
  abstract paintings using non-linear matrix completion,'' in {\em CVPR}, 2016.

\bibitem{deng2009imagenet}
J.~Deng, W.~Dong, R.~Socher, L.-J. Li, K.~Li, and L.~Fei-Fei, ``Imagenet: A
  large-scale hierarchical image database,'' in {\em CVPR}, 2009.

\bibitem{you2016building}
Q.~You, J.~Luo, H.~Jin, and J.~Yang, ``Building a large scale dataset for image
  emotion recognition: The fine print and the benchmark,'' in {\em AAAI}, 2016.

\bibitem{hu2014semantic}
C.~Hu, Z.~Xu, Y.~Liu, L.~Mei, L.~Chen, and X.~Luo, ``Semantic link
  network-based model for organizing multimedia big data,'' {\em IEEE
  Transactions on Emerging Topics in Computing}, vol.~2, no.~3, pp.~376--387,
  2014.

\bibitem{cui2016sentiment}
Z.~Cui, X.~Shi, and Y.~Chen, ``Sentiment analysis via integrating distributed
  representations of variable-length word sequence,'' {\em Neurocomputing},
  vol.~187, pp.~126--132, 2016.

\bibitem{shepstone2014using}
S.~E. Shepstone, Z.-H. Tan, and S.~H. Jensen, ``Using audio-derived affective
  offset to enhance tv recommendation,'' {\em IEEE Transactions on Multimedia},
  vol.~16, no.~7, pp.~1999--2010, 2014.

\bibitem{poria2016fusing}
S.~Poria, E.~Cambria, N.~Howard, G.-B. Huang, and A.~Hussain, ``Fusing audio,
  visual and textual clues for sentiment analysis from multimodal content,''
  {\em Neurocomputing}, vol.~174, pp.~50--59, 2016.

\bibitem{hanjalic2005affective}
A.~Hanjalic and L.-Q. Xu, ``Affective video content representation and
  modeling,'' {\em IEEE Transactions on Multimedia}, vol.~7, no.~1,
  pp.~143--154, 2005.

\bibitem{soleymani2014corpus}
M.~Soleymani, M.~Larson, T.~Pun, and A.~Hanjalic, ``Corpus development for
  affective video indexing,'' {\em IEEE Transactions on Multimedia}, vol.~16,
  no.~4, pp.~1075--1089, 2014.

\bibitem{yadati2014cavva}
K.~Yadati, H.~Katti, and M.~Kankanhalli, ``Cavva: Computational affective
  video-in-video advertising,'' {\em IEEE Transactions on Multimedia}, vol.~16,
  no.~1, pp.~15--23, 2014.

\bibitem{sun2016sentiment}
X.~Sun, C.~Li, and F.~Ren, ``Sentiment analysis for chinese microblog based on
  deep neural networks with convolutional extension features,'' {\em
  Neurocomputing}, vol.~210, pp.~227--236, 2016.

\bibitem{xu2008}
M.~Xu, J.~S. Jin, S.~Luo, and L.~Duan, ``Hierarchical movie affective content
  analysis based on arousal and valence features,'' in {\em ACM MM}, 2008.

\bibitem{benini2011connotative}
S.~Benini, L.~Canini, and R.~Leonardi, ``A connotative space for supporting
  movie affective recommendation,'' {\em IEEE Transactions on Multimedia},
  vol.~13, no.~6, pp.~1356--1370, 2011.

\bibitem{tarvainen2014content}
J.~Tarvainen, M.~Sjoberg, S.~Westman, J.~Laaksonen, and P.~Oittinen,
  ``Content-based prediction of movie style, aesthetics, and affect: Data set
  and baseline experiments,'' {\em IEEE Transactions on Multimedia}, vol.~16,
  no.~8, pp.~2085--2098, 2014.

\bibitem{tang2012quantitative}
J.~Tang, Y.~Zhang, J.~Sun, J.~Rao, W.~Yu, Y.~Chen, and A.~C.~M. Fong,
  ``Quantitative study of individual emotional states in social networks,''
  {\em IEEE Transactions on Affective Computing}, vol.~3, no.~2, pp.~132--144,
  2012.

\bibitem{peng2015mixed}
K.-C. Peng, T.~Chen, A.~Sadovnik, and A.~C. Gallagher, ``A mixed bag of
  emotions: Model, predict, and transfer emotion distributions,'' in {\em
  CVPR}, 2015.

\bibitem{mikels2005emotional}
J.~A. Mikels, B.~L. Fredrickson, G.~R. Larkin, C.~M. Lindberg, S.~J. Maglio,
  and P.~A. Reuter-Lorenz, ``Emotional category data on images from the
  international affective picture system,'' {\em Behavior research methods},
  vol.~37, no.~4, pp.~626--630, 2005.

\bibitem{borth2013large}
D.~Borth, R.~Ji, T.~Chen, T.~Breuel, and S.-F. Chang, ``Large-scale visual
  sentiment ontology and detectors using adjective noun pairs,'' in {\em ACM
  MM}, 2013.

\bibitem{zhao2016predicting}
S.~Zhao, H.~Yao, Y.~Gao, G.~Ding, and T.-S. Chua, ``Predicting personalized
  image emotion perceptions in social networks,'' {\em IEEE Transactions on
  Affective Computing}, 2016.

\bibitem{yanulevskaya2008}
V.~Yanulevskaya, J.~Van~Gemert, K.~Roth, A.-K. Herbold, N.~Sebe, and J.-M.
  Geusebroek, ``Emotional valence categorization using holistic image
  features,'' in {\em ICIP}, 2008.

\bibitem{solli2009}
M.~Solli and R.~Lenz, ``Color based bags-of-emotions,'' in {\em CAIP}, 2009.

\bibitem{lu2012shape}
X.~Lu, P.~Suryanarayan, R.~B. Adams~Jr, J.~Li, M.~G. Newman, and J.~Z. Wang,
  ``On shape and the computability of emotions,'' in {\em ACM MM}, 2012.

\bibitem{chen2014object}
T.~Chen, F.~X. Yu, J.~Chen, Y.~Cui, Y.-Y. Chen, and S.-F. Chang, ``Object-based
  visual sentiment concept analysis and application,'' in {\em ACM MM}, 2014.

\bibitem{tkalcic2013affective}
M.~Tkalcic, A.~Odic, A.~Kosir, and J.~Tasic, ``Affective labeling in a
  content-based recommender system for images,'' {\em IEEE transactions on
  Multimedia}, vol.~15, no.~2, pp.~391--400, 2013.

\bibitem{zeiler2014visualizing}
M.~D. Zeiler and R.~Fergus, ``Visualizing and understanding convolutional
  networks,'' in {\em ECCV}, 2014.

\bibitem{lu2014rapid}
X.~Lu, Z.~Lin, H.~Jin, J.~Yang, and J.~Z. Wang, ``Rapid: rating pictorial
  aesthetics using deep learning,'' in {\em ACM MM}, 2014.

\bibitem{andrearczyk2016using}
V.~Andrearczyk and P.~F. Whelan, ``Using filter banks in convolutional neural
  networks for texture classification,'' {\em Pattern Recognition Letters},
  vol.~84, pp.~63--69, 2016.

\bibitem{szegedy2015going}
C.~Szegedy, W.~Liu, Y.~Jia, P.~Sermanet, S.~Reed, D.~Anguelov, D.~Erhan,
  V.~Vanhoucke, and A.~Rabinovich, ``Going deeper with convolutions,'' in {\em
  CVPR}, 2015.

\bibitem{he2016deep}
K.~He, X.~Zhang, S.~Ren, and J.~Sun, ``Deep residual learning for image
  recognition,'' in {\em CVPR}, pp.~770--778, 2016.

\bibitem{yanulevskaya2012eye}
V.~Yanulevskaya, J.~Uijlings, E.~Bruni, A.~Sartori, E.~Zamboni, F.~Bacci,
  D.~Melcher, and N.~Sebe, ``In the eye of the beholder: employing statistical
  analysis and eye tracking for analyzing abstract paintings,'' in {\em ACM
  MM}, 2012.

\bibitem{lang2008international}
P.~J. Lang, M.~M. Bradley, and B.~N. Cuthbert, ``International affective
  picture system (iaps): Affective ratings of pictures and instruction
  manual,'' {\em Technical report A-8}, 2008.

\bibitem{sartori2015s}
A.~Sartori, D.~Culibrk, Y.~Yan, and N.~Sebe, ``Who's afraid of itten: Using the
  art theory of color combination to analyze emotions in abstract paintings,''
  in {\em ACM MM}, 2015.

\bibitem{Simonyan14c}
K.~Simonyan and A.~Zisserman, ``Very deep convolutional networks for
  large-scale image recognition,'' {\em CoRR}, vol.~abs/1409.1556, 2014.

\bibitem{joachims1999transductive}
T.~Joachims, ``Transductive inference for text classification using support
  vector machines,'' in {\em ICML}, 1999.

\bibitem{chen2015matrix}
C.-H. Chen, V.~M. Patel, and R.~Chellappa, ``Matrix completion for resolving
  label ambiguity,'' in {\em CVPR}, 2015.

\end{thebibliography}
\end{document}